\documentclass[runningheads,a4paper]{llncs}

\usepackage{amssymb}
\usepackage{graphicx}
\usepackage{amsmath}

\usepackage[hyphens]{url}
\urldef{\mailsa}\path|abc@example.com|
\newcommand{\keywords}[1]{\par\addvspace\baselineskip
\noindent\keywordname\enspace\ignorespaces#1}

\usepackage{tabu}
\usepackage{hyperref}
\usepackage{subfig}
\usepackage{todonotes}

\usepackage{algorithm}
\usepackage{algorithmic}

\usepackage{adjustbox}
\usepackage{alltt}
\usepackage{pmboxdraw}

\newcommand{\citet}[1]{\citeauthor{#1} \shortcite{#1}}

\newcommand{\titletext}{Handling Collocations in Hierarchical Latent Tree Analysis for Topic Modeling}

\renewcommand{\vec}[1]{\boldsymbol#1}
\newcommand{\word}[1]{\texttt{#1}}
\newcommand{\sysname}[1]{\textsc{#1}}

\definecolor{darkblue}{rgb}{0,0,.8}

\begin{document}

\mainmatter  

\title{Handling Collocations in Hierarchical Latent Tree Analysis for Topic Modeling}

\titlerunning{Handling Collocations in HLTA for Hierarchical Topic Modeling}

%
%
\author{Leonard K. M. Poon\inst{1} \and Nevin L. Zhang\inst{2} \and Haoran Xie\inst{1} \and Gary Cheng\inst{1}}
\authorrunning{Anonymous Authors}

\institute{Department of Mathematics and Information Technology\\
The Education University of Hong Kong, Hong Kong SAR \\
\email{kmpoon@eduhk.hk}, \email{hxie@eduhk.hk}, \email{chengks@eduhk.hk}
\and Department of Computer Science and Engineering\\
The Hong Kong University of Science and Technology, Hong Kong SAR\\
\email{lzhang@cse.ust.hk}
}
%
%

\toctitle{\titletext}
\tocauthor{Anonymous authors}
\maketitle

\begin{abstract}

Topic modeling has been one of the most active research areas in machine learning in recent years.  Hierarchical latent tree analysis (HLTA) has been recently proposed for hierarchical topic modeling and has shown superior performance over state-of-the-art methods.  However, the models used in HLTA have a tree structure and cannot represent the different meanings of multiword expressions sharing the same word appropriately.  Therefore, we propose a method for extracting and selecting collocations as a preprocessing step for HLTA.  The selected collocations are  replaced with single tokens in the bag-of-words model  before running HLTA.  Our empirical evaluation shows that the proposed method led to better performance of HLTA on three of the four data sets tested.

\keywords{Collocations, Topic modeling, Hierarchical latent tree analysis, Latent tree models, Document clustering}
\end{abstract}

\section{Introduction}

Topic modeling has been one of the most active research areas in machine learning in recent years.  Most methods for topic modeling are based on \emph{latent Dirichlet allocation} (LDA)~\cite{Blei2003aa}.  The basic version of LDA yields a flat level of topics.  It has later been extended by nested Chinese restaurant process (nCRP)~\cite{Blei2010aa} and nested hierarchical Dirichlet processes (nHDP)~\cite{Paisley2015aa} to produce topic hierarchies with multiple levels of topics.  Recently, Chen et al.~\cite{Chen2017aa} have proposed another approach, called hierarchical latent tree analysis (HLTA), for hierarchical topic modeling.  They have shown that HLTA produces topic hierarchies of higher quality than the two state-of-the-art LDA-based methods nCRP and nHDP.

In HLTA, words in the vocabulary are represented by observed variables and
topics by latent variables.  The word variables are connected to the topic variables to form a tree-structured probabilistic model.  Due to the tree structure, a word variable can appear as the leaf in only one branch of the topic hierarchy.  This presents a difficulty for representing the different meanings of those multiword expressions that contain the same word.  For example, the word ``network'' can appear in multiword expressions such as ``neural network'', ``Bayesian network'', ``Markov network'', ``social network'', etc.  It is more reasonable if those terms appear separately in different branches of a topic hierarchy.

Following \cite{Sag2002aa}, we define \emph{multiword expressions} as idiosyncratic interpretations that cross word boundaries and  \emph{collocations} as sequences of words with statistically significant co-occurrences.  Note that collocations include all forms of multiword expressions,  but they also include frequently occurring phrases that may not be considered as multiword expressions.  Lau et al.~\cite{Lau2013aa} showed the quality of topic models produced by LDA can be improved by replacing collocations with single tokens.  We adopt that preprocessing approach to mitigate the limitation induced by the tree structure of models used in HLTA.

In this paper, we propose a method for extracting and selecting collocations for hierarchical topic modeling.  The selected collocations are replaced with single tokens in the bag-of-words model before running HLTA. The proposed method can automatically determine the number of collocations to be selected and can find collocations with more than two words.  It is run in the preprocessing phase and does not induce any additional complexity to the models used in HLTA.

In the following, we review background of our work in Section~\ref{sec:background} and discuss related work in Section~\ref{sec:related}.  Next, the proposed method is explained and  empirically evaluated in Sections~\ref{sec:collocations} and \ref{sec:evaluation}.  Finally, Section~\ref{sec:conclusions} concludes our work.

\section{Background} \label{sec:background}


Consider a collection $\mathcal{D}=\{d_1, \ldots, d_N\}$ of $N$ documents.   Suppose $P$ words are used in the vocabulary $\mathcal{V}=\{w_1, \ldots, w_P\}$.  Each document $d$ can be represented by the bag-of-words model using a vector $d=(c_1, \ldots, c_P)$, where $c_i$ denotes the count of word $w_i$ occurring in document $d$.  Topic modeling aims to detect a number $K$ of topics $z_1, \ldots, z_K$ among  documents $\mathcal{D}$, where $K$ can be given or learned from data.  A topic is often characterized by representative words according to the distribution given by the model.

%

\begin{figure}[tbp]
\centering
\includegraphics[width=.9\textwidth]{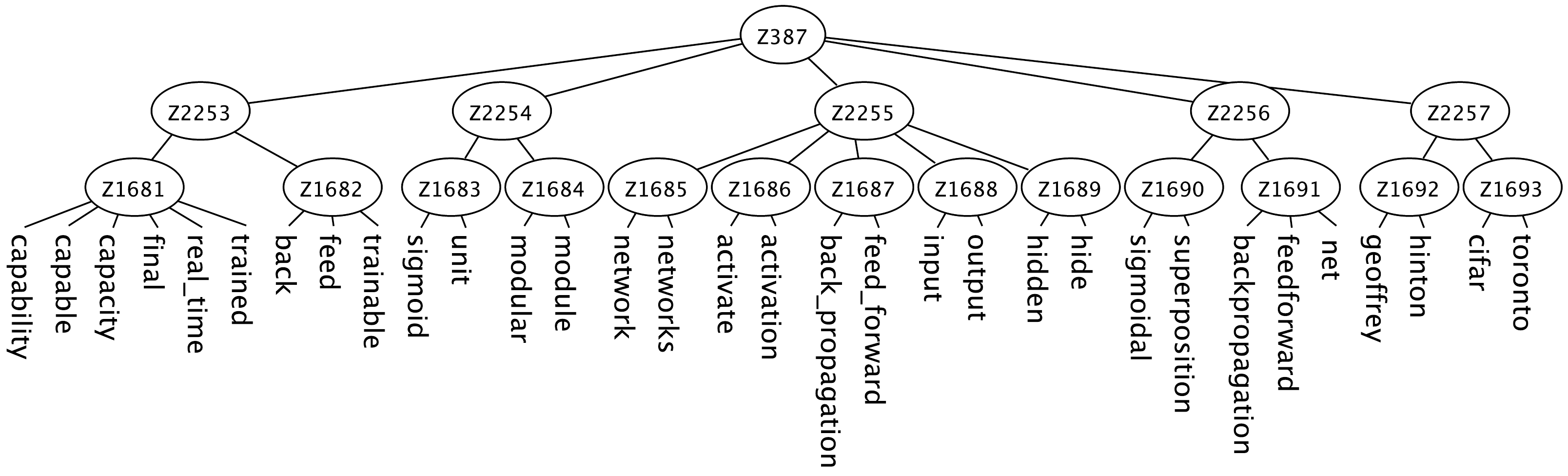} \vspace{-2ex}
\caption{Subtree of a latent tree model obtained by HLTA on the NIPS data set.  The oval nodes represent topic variables whereas the text nodes without borders represent word variables. All variables have two states.} \label{fig:subtree}
\vspace{-1em}
\end{figure}

\emph{Hierarchical latent tree analysis} (HLTA) is a recently proposed method for hierarchical topic modeling~\cite{Chen2017aa}.  It is based on a class of tree-structured models known as \emph{latent tree models} (LTMs)~\cite{Chen2012ab,Zhang2004aa}.  Fig.~\ref{fig:subtree} shows an example of LTM.  When an LTM is used for topic modeling, the leaf nodes represent the observed word variables $\vec{W}$, whereas the internal nodes represent the unobserved topic variables $\vec{Z}$.   All variables are binary.  Each word variable $W_i \in \vec{W}$ indicates the presence or absence of the word $w_i \in \mathcal{V}$ in a document.  Each topic variable $Z_i \in \vec{Z}$ indicates whether a document belongs to the $i$-th topic.

For technical convenience, an LTM is often rooted at one of its latent nodes.  It can then be regarded as a Bayesian network
 with edges directed away from the root. The numerical information of the model includes a marginal distribution for the root and one conditional distribution for each edge. For example, edge $Z1685$ $\rightarrow$ \word{network} in Fig.~\ref{fig:subtree} is associated with probability $P(\word{network} | Z1685)$. The conditional distribution associated with each edge characterizes the probabilistic dependence between the two nodes that the edge connects. The product of all those distributions defines a joint distribution over all the latent variables $\vec{Z}$ and observed variables $\vec{W}$.  Denote the parent of a variable $X$ as $pa(X)$ and let $pa(X)$ be an empty set when $X$ is the root. Then the LTM defines a joint distribution over all observed and latent variables as follows:
$$P(\vec{W},\vec{Z}) = \prod_{X\in \vec{W}\cup\vec{Z}} P(X|pa(X)).$$

Given a document $d$, the values of the binary word variables $\vec{W}$ are observed.  We use $d = (w_1, \ldots,w_P)$ to denote also those observed values.  Whether a document $d$ belongs to a topic $Z \in \vec{Z}$ can be determined by the probability $P(Z|d)$.  The LTM allows each document to belong to multiple topics.

For topic modeling, an LTM has to be learned from the document data $\mathcal{D}$.  We use the \sysname{PEM-HLTA} method~\cite{Chen2017aa} to build LTMs for topic modeling.  The method builds LTMs level by level and is thus also known as hierarchical latent tree analysis. Intuitively, the co-occurrence of words are captured by the latent variables at the lowest level, whose co-occurring patterns are captured by latent variables at higher levels. In the topic hierarchy, the topics at higher levels are more general and those at lower levels are more specific.

To extract the topic hierarchy from an LTM, we follow the tree structure in the model and use each internal node to represent a topic.  A topic is characterized by the words most relevant to them.  Specifically, we compute the mutual information (MI)~\cite{Cover2006aa} between a topic variable and each of its descendent word variable.  Then we pick at most seven descendent words with the highest MI to characterize the topic.

\section{Related Work} \label{sec:related}

The bag-of-words model allows efficient computation but does not preserve word order.  To make use of word order for better performance, topic models have been extended such that collocations can be generated in a unified model.  Extensions of topic models have been proposed for LDA~\cite{Wallach2006aa,Griffiths2007aa,Wang2007aa} and Hierarchical Dirichlet Processes~\cite{Jameel2015aa}.

%
Lau et al.~\cite{Lau2013aa} have pointed out a drawback of unified models is that they in general induce additional computational overhead and model complexity.  Lau et al.~\cite{Lau2013aa} instead extracted collocations in a preprocessing phase.  The collocations were then represented as single tokens in the bag-of-words model before learning a standard topic model.  They used the N-gram Statistics Package (NSP)~\cite{Banerjee2003aa} for extracting collocations and used LDA for topic modeling.  Only bigrams were considered in their study.  Nokel and Loukachevitch~\cite{Nokel2016aa} followed the collocation extraction approach.  They modified the parameter estimation method LDA to such that bigrams and unigrams belong to the same topics more often.

Our work follows the collocation extraction approach.  However, we use HLTA for topic modeling.  We also propose an algorithm for extracting collocation using TF-IDF as selection criterion.  The proposed algorithm allows us to automatically determine the number of collocations and include collocations of any length.

HLTA was first proposed by Liu et al.~\cite{Liu2014ab} and a method for scaling up was later proposed by Chen et al.~\cite{Chen2016ab}.  The fundamental differences between HLTA and LDA-based methods were discussed in~\cite{Chen2017aa}.  HLTA has been applied on textual reviews~\cite{Poon2017af} and for building a topic browsing system~\cite{Poon2017aa}.  Collocations have been briefly considered in HLTA~\cite{Chen2017aa,Poon2017aa}. However, those studies did not describe the method in details and did not show any quantitative evaluation.

\section{Handling Collocations} \label{sec:collocations}

Fig.~\ref{fig:subtree} shows part of the LTM obtained by HLTA on a collection of NIPS papers.  It can be used to illustrate the limitation of LTMs when dealing with collocations.  The subtree is obviously related to neural networks and contains \word{network} as one of its leaf nodes.  Due to the tree structure, the node \word{network} cannot be connected to other subtrees related to Bayesian networks or social networks.  It explains why standard HLTA cannot handle collocations appropriately.

\begin{algorithm}[tbp]
\caption{\texttt{Preprocess}$(\mathcal{D}, r, P)$}
\label{algo:preprocess}

\textbf {Input}:  $\mathcal{D}$ -- Document collection, $r$ -- number of concatentations, $P$ -- number of tokens to be selected.\\
\textbf {Output}: $\mathcal{D}$ -- Document collection with individual words replaced with selected collocations, $\mathcal{V}$ -- Vocabulary with selected tokens (individual words and collocations). \\
\vspace{-0.5em}

\begin{algorithmic}[1]
\STATE Compute TF-IDF of tokens (individual words) in $\mathcal{D}$.  Set $\mathcal{V}$ to be the set of $P$ tokens with highest T-IDF values.
\STATE If $r = 0$, {\bf return} $\mathcal{D}$ and $\mathcal{V}$.
\FOR{$i = 1$ to $r$}
\STATE For every pair of consecutive tokens $t_1$ and $t_2$ in $\mathcal{D}$, form a new token $t'$ by concatenation if $t_1,t_2 \in \mathcal{V}$.  
\STATE Compute TF-IDF of the concatenated tokens and the original tokens in $\mathcal{D}$.  Set $\mathcal{V}$ to be the set of $P$ tokens with highest TF-IDF values.
\STATE For every pair of consecutive tokens $t_1$ and $t_2$ occurred in $\mathcal{D}$, replace the pair with the concatenated token $t'$ if $t' \in \mathcal{V}$.
\ENDFOR
\STATE Compute TF-IDF of the tokens (including concatenated tokens and individual words) in $\mathcal{D}$.  Set $\mathcal{V}$ to be set of the $P$ tokens with highest TF-IDF values.
\STATE {\bf return} $\mathcal{D}$ and $\mathcal{V}$.
\end{algorithmic}
\end{algorithm}

To handle collocations in HLTA, we extract collocations from a document collection $\mathcal{D}$ in a preprocessing phrase.  The algorithm for preprocessing  is given in Algorithm~\ref{algo:preprocess}.  To explain the algorithm, we assume $\mathcal{D}$ contains the collocation ``neural network'' and show how the algorithm can extract that collocation with number of concatenations $r$ set to 1.

The algorithm starts by treating individual words as tokens (Line 1).  It computes their TF-IDF values as given below:
$$\textrm{tf-idf}(t) = \frac{1}{\ln \textrm{df}(t)} \sum_{d\in\mathcal{D}}\textrm{tf}(t,d),$$
where the term frequency $\textrm{tf}(t,d)$ is defined as the number of occurrences of a token $t$ in document $d$, and the document frequency $\textrm{df}(t)$ is defined as the number of documents that contain the token $t$.  

During the first iteration, the algorithm considers every pair of consecutive words (e.g. \word{neural} and \word{network}).  If both words are contained in the current vocabulary $\mathcal{V}$, then a new token (e.g. \word{neural-network}) is formed (Line 6).  Next, the TF-IDF values of the new tokens and the original tokens are computed (Line 7).  After that, vocabulary $\mathcal{V}$ is updated by selecting tokens among the new tokens and original tokens.  If a new token such as \word{network-network} is selected in $\mathcal{V}$, every pair of the component tokens will be replaced with the new token.

After the iteration, TF-IDF is computed again to update vocabulary $\mathcal{V}$ (Line 10).  It is because the TF-IDF values of the component tokens will be affected after being replaced by the newly formed tokens.  The updated vocabulary and the updated document collection are then returned by the algorithm.  They will be used to build bag-of-words representation of the updated document collection as input to HLTA.  In HLTA, the leaf nodes of LTMs represent word variables.  We keep using the term word variables for brevity even though they may represent individual words or tokens formed by concatenation.

The number of concatenations $r$ can be larger than 1 if longer collocations should be considered.  Note that the maximum length of collocations that can be considered by the algorithm is $2^{r}$.  When $r=0$, the algorithm performs standard preprocessing step for HLTA considering only individual words.


\begin{table}[tbp]
\caption{Properties of data sets.  The table shows the number of documents ($D$), the average document length ($L$), the number of distinct words ($V$), and the number of collocations selected by \texttt{Preprocess} when $P=5000$ and $r=1$.} \label{table:data-sets}
\centering
\setlength{\tabcolsep}{1em}

\begin{tabular}{ccccccc} \hline
 & $D$ & $L$ & $V$ & \# Collocations \\ \hline
NIPS & 7,241 & 1,988 & 275,344 & 1,224\\
AAN & 23,380 & 2,256 & 1,204,640 & 969\\ 
JRC & 23,545 & 627 & 173,700 & 1,706 \\ 
Reuters & 19,043 & 74 & 43,726 & 1,408\\
\end{tabular}

\vspace{-1em}
\end{table}

\section{Empirical Evaluation} \label{sec:evaluation}

Four text collections were used in our experiments (see Table~\ref{table:data-sets}).  The NIPS collection contains papers published during 1987-2016.\footnote{\url{https://www.kaggle.com/benhamner/nips-papers}}  The AAN corpus~\cite{Radev2013aa} contains papers on natural language processing from the ACL Anthology Network. 
The JRC collection contains European Union documents of mostly legal nature from the English part of the JRC-Acquis corpus~\cite{Ralf2006aa}.
The Reuters-21578 collection contains documents appeared on the Reuters newswire in 1987.\footnote{\url{http://archive.ics.uci.edu/ml/datasets/Reuters-21578+Text+Categorization+Collection}}  
We used underscore to replace all non-alphanumeric characters.  We removed stop words
and words shorter than 3 characters.  We used Stanford CoreNLP~\cite{Manning2014aa} for lemmatization.

We set the size of vocabulary $P$ to 5,000 for all data sets.  Our proposed method (HLTA-r1) ran \texttt{Preprocess} with $r=1$ and then HLTA.  Two other methods were included for comparison.  The first one was HLTA without considering collocations.  The second one followed Lau et al.~\cite{Lau2013aa} and used NSP~\cite{Banerjee2003aa} to extract bigrams based on the Student’s t-test.  The number of bigrams was set to 1,000 since it was shown to have the best performance in \cite{Lau2013aa}.  

\begin{table}[tbp]
\caption{Topic coherence attained by various methods.  The averages and standard deviations for 10 runs are reported. Best scores are marked in bold.  Our proposed method (HLTA-r1) performed best on 3 out of 4 data sets.} \label{table:topic-coherence}
\centering
\setlength{\tabcolsep}{.8em}

\begin{tabular}{llllll} \hline
& & \multicolumn{1}{c}{NIPS} & \multicolumn{1}{c}{AAN} & \multicolumn{1}{c}{JRC} & \multicolumn{1}{c}{Reuters} \\ \hline
HLTA & base & -8.82$\pm$0.06 & -8.67$\pm$0.07 & -9.18$\pm$0.06 & -13.25$\pm$0.18 \\
& NSP & -8.48$\pm$0.08 & -8.62$\pm$0.05 & -8.72$\pm$0.05 & -13.34$\pm$0.14 \\
& r1 & \textbf{-8.42$\pm$0.07} & \textbf{-8.52$\pm$0.04} & \textbf{-8.54$\pm$0.06} & -13.26$\pm$0.11 \\ \hline
nHDP & base & -13.16$\pm$0.07 & -12.68$\pm$0.14 & -10.09$\pm$0.18 & -13.45$\pm$0.24 \\
 & NSP & -12.80$\pm$0.14 & -12.44$\pm$0.18 & -10.78$\pm$0.17 & -14.21$\pm$0.20 \\
 & r1 & -12.75$\pm$0.12 & -12.68$\pm$0.14 & -10.09$\pm$0.18 & -13.45$\pm$0.24 \\ \hline
nCPR & base & -9.47$\pm$0.06 & \multicolumn{1}{c}{--} & \multicolumn{1}{c}{--} & \textbf{-12.10$\pm$0.13} \\
 & NSP & -9.52$\pm$0.13 & \multicolumn{1}{c}{--} & \multicolumn{1}{c}{--} & -13.13$\pm$0.19 \\
 & r1 & -9.54$\pm$0.01 & \multicolumn{1}{c}{--} & \multicolumn{1}{c}{--} & -13.38$\pm$0.16 \\ \hline
\end{tabular}

\vspace{-1em}
\end{table}

We measure topic quality using the topic coherence score~\cite{Mimno2011aa}: 
$$TC(\vec{w}) = \sum_{i=2}^{M}\sum_{j=1}^{i-1}\frac{df(w_{i}, w_{j})+1}{df(w_{j})},$$
where $\vec{w}$ are the top-M words characterizing a topic, $df(w_{i}, w_{j}$ is the number of documents containing both words $w_{i}$ and $w_{j}$, and $df(w_{j})$ is the number of documents containing word $w_{j}$.  We used $M=4$ following~\cite{Chen2017aa}.  The score of a method is given by the average of those topics with at least four keywords.

Table~\ref{table:topic-coherence} shows the topic coherence scores.  On the NIPS, AAN, JRC data sets, HLTA produced better topics when collocations were considered.  Higher scores were obtained by our proposed method HLTA-r1 than by HLTA-NSP.  It shows that our proposed algorithm \texttt{Preprocess} is more effective in extracting and selecting collocations than NSP.  One possible reason is that \texttt{Preprocess} may be able to find a more appropriate number of collocations.   Table~\ref{table:data-sets} shows that the number of collocations selected by \texttt{Preprocess} varied in different data sets.  On the Reuters data set, the standard HLTA performed slightly better than HLTA-r1.  It showed that considering collocations may not help improve topic modeling on this data, possibly because of the short documents.  However, it did not worsen materially when considering collocations using our proposed method.

\begin{table}[tbp]
\caption{Some of the topics related to collocations with word \word{network} obtained  by our proposed method on the NIPS data set along with their parent topics.} \label{table:hierarchy} \vspace{-1em}
\begin{alltt} \scriptsize
graph edge node vertex undirected graph-node undirected-graph
\textSFii\textSFx\textSFx \textcolor{darkblue}{social-network} social relational kleinberg entity link-prediction friend \smallskip
belief-propagation loopy-belief-propagation markov-random-field mrf pearl partition-function loopy
\textSFii\textSFx\textSFx pearl reasoning \textcolor{darkblue}{belief-network} frey causal-inference causal identifiability\smallskip
inference graphical-model infer probabilistic probabilistic-model approximate-inference marginal
\textSFii\textSFx\textSFx probabilistic probabilistic-model marginal joint-distribution joint \textcolor{darkblue}{bayesian-network} dependency\smallskip
deep deep-learning deep-convolutional sutskever \textcolor{darkblue}{deep-neural-network} \textcolor{darkblue}{deep-network} krizhevsky
\textSFviii\textSFx\textSFx deep deep-learning \textcolor{darkblue}{deep-neural-network} \textcolor{darkblue}{deep-network} bengio lecun salakhutdinov
\textSFviii\textSFx\textSFx deep-convolutional sutskever krizhevsky convolutional \textcolor{darkblue}{convolutional-neural-network} imagenet cnn
\textSFii\textSFx\textSFx courville goodfellow deep-belief generative-adversarial restricted-boltzmann-machine gan
    \textcolor{darkblue}{generative-adversarial-network}
\end{alltt}\vspace{-2em}
\end{table}

Table~\ref{table:hierarchy} shows some of the topics containing collocations with the word \word{network} and their parent topics obtained by HLTA on the NIPS data set.  The collocations were extracted by running \texttt{Preprocess} with $r=2$. We see that those collocations with \word{network} now appear in different branches of the topic hierarchy with more related parents.  It shows how the limitation of LTMs for representing different meanings of collocations can be mitigated.  Besides, we see  that some meaningful collocations with three words (e.g. \word{deep-neural-network}, \word{convolutional-neural-network}, and \word{generative-adversarial-network}) were found.  It shows that our proposed algorithm \texttt{Preprocess} can be effective in finding longer collocations.

\section{Conclusions} \label{sec:conclusions}

We propose a method for extracting and selecting collocations based on TF-IDF.  The selected collocations are replaced with single tokens in the bag-of-words model before running HLTA.  The proposed method can automatically determines the number of collocations to be selected and can find collocations with more than two words.   Our experiments show that the proposed method in general leads to better performance of HLTA.  An implementation of our proposed method can be found online.\footnote{URL is withheld for anonymity.}

%

\bibliographystyle{splncs03}
\bibliography{papers}

\end{document}